\pdfoutput=1
\documentclass{llncs}
\usepackage{bbding}
\usepackage{cite}
\usepackage{amsmath,amssymb,amsfonts}
\usepackage{algorithmic}
\usepackage{graphicx}
\usepackage{textcomp}
\usepackage{xcolor}
\usepackage{epsfig}
\usepackage{graphicx}
\usepackage{multirow}
\usepackage{soul}
\usepackage{subcaption}
\usepackage{caption}
\usepackage[font=small,labelfont=bf]{caption}

\usepackage{algorithm2e}
\usepackage{soul}
\usepackage{color,xcolor}
\usepackage{makeidx}  
\begin{document}
\frontmatter          
\pagestyle{headings}  
\addtocmark{BIPA} 
\mainmatter              
\title{Economical Precise Manipulation and Auto Eye-Hand Coordination with Binocular Visual Reinforcement Learning}
\titlerunning{BIPA}  
%
\author{Yiwen Chen, Sheng Guo, Zedong Zhang, Lei Zhou, Xian Yao Ng, Marcelo H. Ang, Jr.}
\authorrunning{Yiwen Chen, Sheng Guo, Zedong Zhang, Lei Zhou, Xian Yao Ng, Marcelo H. Ang, Jr.} 
%
\tocauthor{Yiwen Chen, Sheng Guo, Zedong Zhang, Lei Zhou, Xian Yao Ng, Marcelo H. Ang, Jr.}
\institute{Advanced Robotics Center, National University of Singapore, Singapore\\
\email{yiwen.chen@u.nus.edu}
}

\maketitle              

\begin{abstract}
Precision robotic manipulation tasks (insertion, screwing, precisely pick, precisely place) are required in many scenarios. Previous methods achieved good performance on such manipulation tasks. However, such methods typically require tedious calibration or expensive sensors. 3D/RGB-D cameras and torque/force sensors add to the cost of the robotic application and may not always be economical. In this work, we aim to solve these but using only weak-calibrated and low-cost webcams. We propose Binocular Alignment Learning (BAL), which could automatically learn the eye-hand coordination and points alignment capabilities to solve the four tasks. Our work focuses on working with unknown eye-hand coordination and proposes different ways of performing eye-in-hand camera calibration automatically. The algorithm was trained in simulation and used a practical pipeline to achieve sim2real and test it on the real robot. Our method achieves a competitively good result with minimal cost on the four tasks. 

\end{abstract}

\section{Introduction\label{sec_intro}}

\begin{figure}
     \centering
     \begin{subfigure}[b]{0.48\linewidth}
         \centering
         \includegraphics[width=0.8\linewidth]{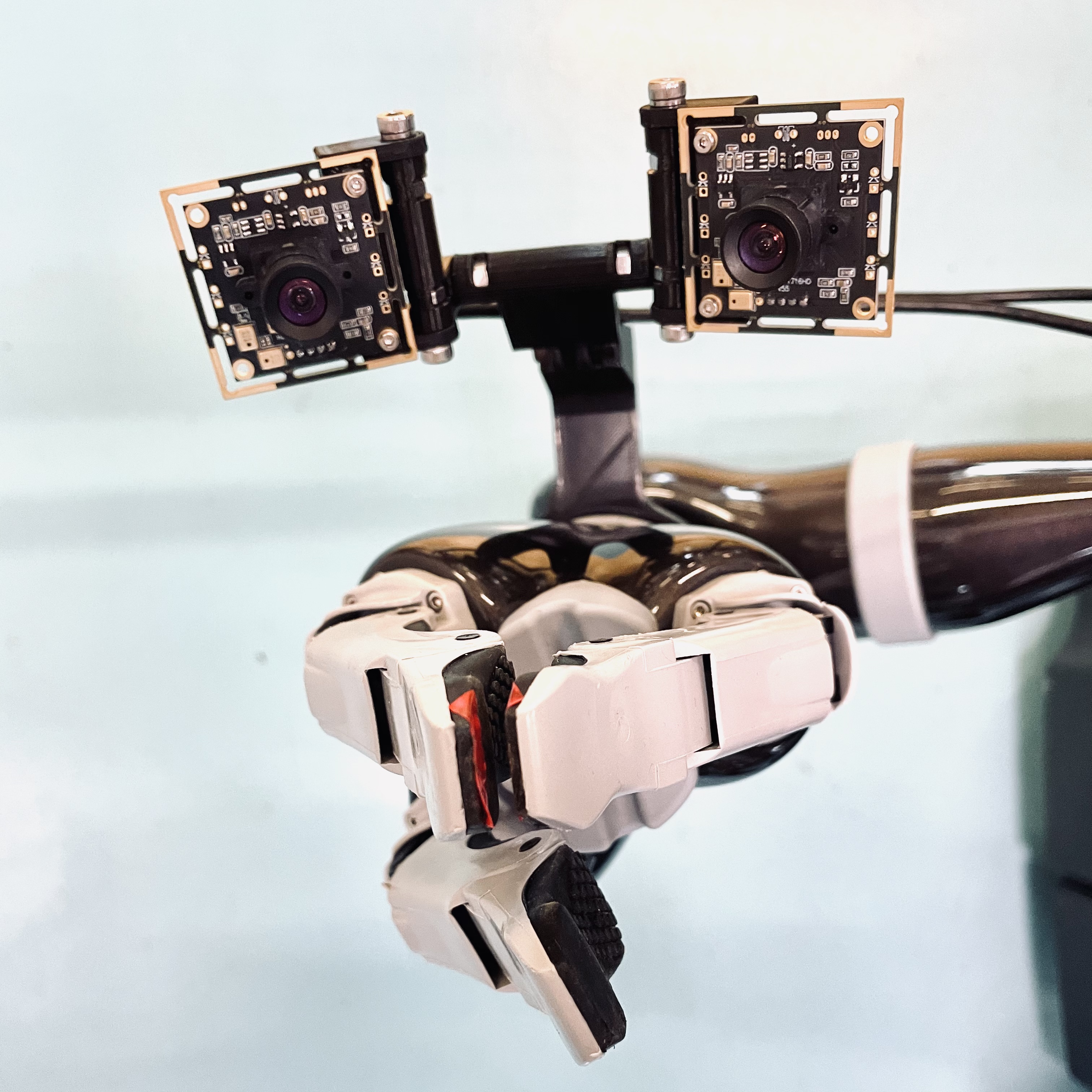}
         \caption{Our Eye-hand System}
         \label{1Bi_CameraGripperSystem}
     \end{subfigure}
     \hfill
     \begin{subfigure}[b]{0.48\linewidth}
         \centering
         \includegraphics[width=0.8\linewidth]{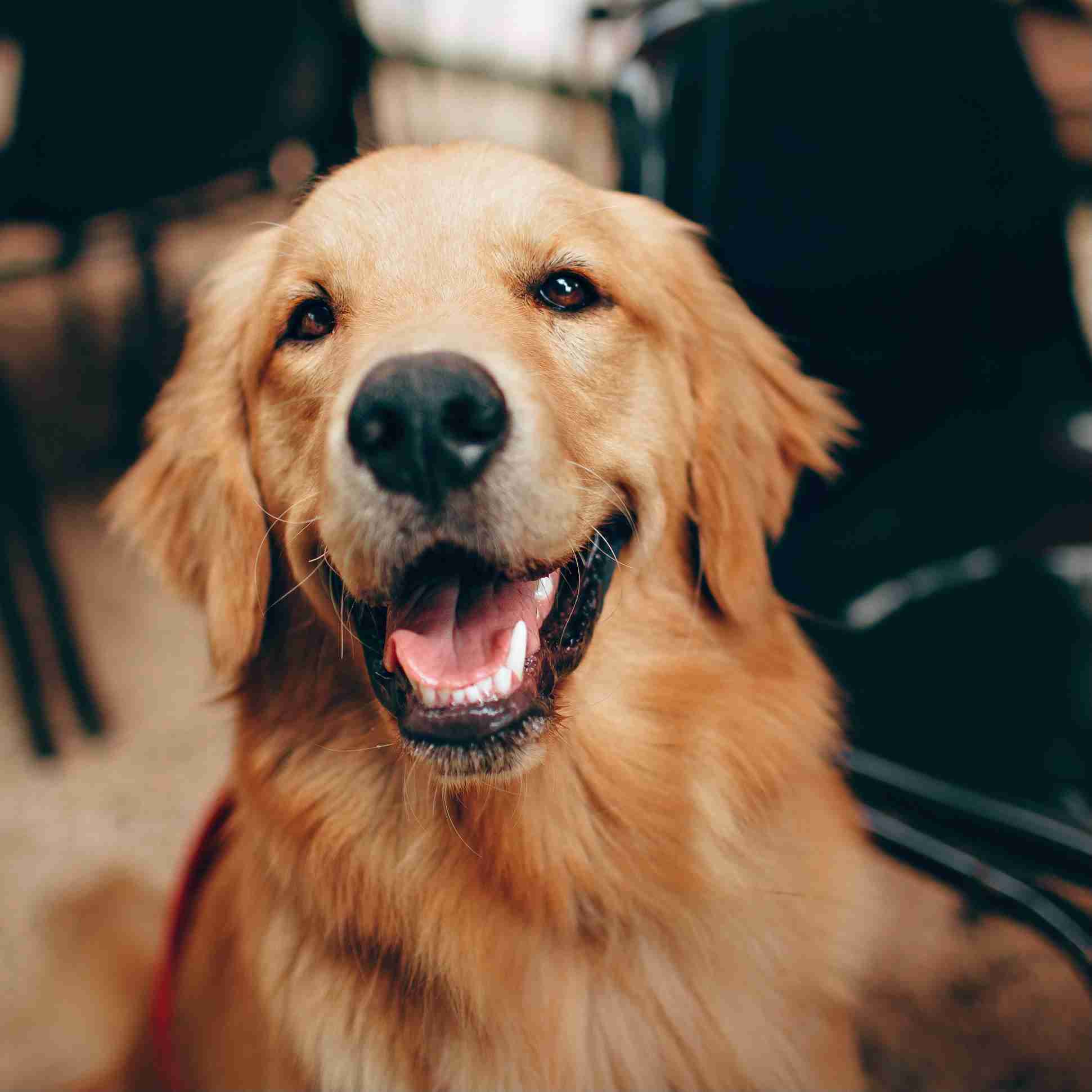}
         \caption{Bio. Eye-Gripper(mouth)}
         \label{1Bi_CameraGripperDogSystem}
     \end{subfigure}
     \hfill
        \caption{Binocular-gripper setup is inspired from biological dual-eyes-one-mouth system.}
        \label{1BinocularImages}
\end{figure}

Precise manipulation is a long-term challenge in robotics area, and it has been seen in a lot of scenarios like gearbox assembly \cite{11prcesision_assembly}, insertion \cite{2_pegoHoleInsert} or small item pick-and-place \cite{Fang_2020_CVPR_graspnet}. 
However, such solutions generally require complex setup, such as the top-down view camera\cite{zeng2019tossingbot}\cite{goodrobot_topdownview_grasp}, high-precision 3D vision camera with dedicate tuning position\cite{12metarl_insert}. Such equipment also lead to higher setup costs. To quantitatively illustrate this, we have listed a cost comparison of those benchmark approaches in Table.\ref{Benchmark}.

We also observe that the human doesn't need a high precision visual system to perform tasks, with our eyes analogous to two RGB cameras. Therefore, in this work, we argue that high precise manipulation tasks can also be done with only visual inputs.

\label{sec_intro_no_eyehand_cali} In the previous influential works, such as TossingBot \cite{zeng2019tossingbot} and 6-DOF GraspNet \cite{graspnet}, visual inputs requires a well calibration. Eye-hand calibration helps increasing task precision. However, over-reliance on eye-hand calibration can lead to a fragile system due to camera position disturbance, camera setup offset, field-of-view changes, setup error, camera support degradation, and so on. The most related paper \cite{eye-hand-cali} also investigate this issue with a learning-based method. Here we address the importance of reducing eye-hand calibration effort and propose our method to challenge performing a precise task under a weak-calibrated eye-hand system. To intentionally introduce eye-hand error, we perform all tasks with the adjustable camera frame.

Our contribution in this work is summarised as follow:

1. Compared with benchmarks, our proposed binocular alignment learning method shows a competitive success rate in insertion $(91.9\pm1.9\%)$, screwing $(93.7\pm6.3\%)$, pick-small-item $(100\%)$, place-small-box $(100\%)$ with the lowest effort and economic cost. And We propose a novel dual-arm working setup in the screwing task, in which the right arm holds the camera and the left arm perform the task.

2. We address the eye-hand calibration issue and propose different auto self-calibration methods SAMLs. We give detailed ablation studies on each SAML method. This solves the unknown eye-hand coordination issue in the tasks.

3. We propose the camera pose randomization training and successfully adapt the learned policy from simulation to the real task (Sim2Real) using domain randomization and feature disentanglement.

\section{Related Work}

For precision manipulation tasks, there are visual-based methods \cite{5_rl_insert}\cite{7_learning_to_touch}\cite{goodrobot_topdownview_grasp}, force-based methods \cite{3_learn_peg_hole_insert} and hybrid methods \cite{1_practial_insert}. In this work, we only look into learning-based and visual-based methods, and select insertion as the key task to research. \cite{1_practial_insert}\cite{5_rl_insert} proposes novel peg-insertion methods using visual DeepRL method. \cite{8_assemblycad}\cite{7_learning_to_touch} proposes an novel pure visual-based solution. However, some additional sensors, information, efforts or costs are required by these methods, such as well camera calibration \cite{1_practial_insert}\cite{8_assemblycad}, high quality visual system\cite{8_assemblycad}, torque/force sensors \cite{1_practial_insert}, hard-to-get information like goal-pose \cite{12metarl_insert}\cite{2_pegoHoleInsert} and goal-pose-image \cite{5_rl_insert}. \cite{eye-hand-cali} proposes a novel way to learn eye-hand calibration using CNN, yet not support camera out of calibrated position.

Reinforcement Learning (RL) has been widely used to solve robotics tasks, such as solving a Rubik's cubic with an anthropomorphic robotic hand \cite{openai2019solvingcubic}, tossing novel objects \cite{zeng2019tossingbot}, performing peg-in-hole insertions \cite{1_practial_insert}\cite{5_rl_insert}\cite{12metarl_insert}. In this work, we also follow Proximal Policy Optimization \cite{ppopaper} to learn the control policy and overcome the camera position disturbance. For the sim2real, there proposed novel methods as domain randomization and augmentation \cite{openai2019solvingcubic}.

Therefore, in this work, we target to propose a learning-based method to achieve high precise tasks, at the same time minimize the economical cost and calibration efforts and perform sim2real.

\section{Preliminaries\label{sec_taskdef}}

\begin{figure}
    \centering
    \includegraphics[width=0.8\linewidth]{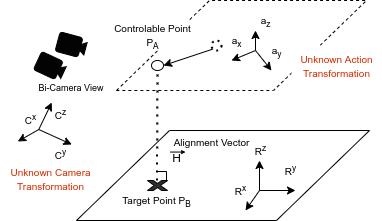}
    \caption{Task Definition}
    \label{fig_task_definition}
\end{figure}

In this work, we model the four tasks insertion, screwing, pick ,and place as Points Alignment tasks. In this 3D task, there are controlled point \(P_A\), target point \(P_B\) and target vector \( \overrightarrow{H}\). The agent should control the position of controllable point \(P_A\) to align the two points \(P_A\) and \(P_B\) with the \(\overrightarrow{H}\), seeing Fig.\ref{fig_task_definition}. This model of task needs the human effort to input the target line  \(\overrightarrow{H}\). We clarify it as user input information shown in Table.\ref{Benchmark} because the PA task will always need a goal. And we also consider it as weak calibration if it's in the manipulation task. But in future work, this target line will be given with using auto line detection without human intervention.

High precision is always required in these tasks. Otherwise, the task will fail. And the setup details can be found in Sec.\ref{sec_experiments} and Fig.\ref{image_task_setup_physical}. To address the weak eye-hand calibration, we assume eye-hand transformation is unknown. To generate unpredictable camera pose errors, we design our camera frame pose adjustable. 

\begin{figure}
     \centering
     \begin{subfigure}{0.23\linewidth}
         \centering
         \includegraphics[width=\linewidth]{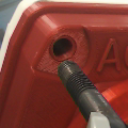}
         \caption{Insert(L)}
         \label{img_insert1}
     \end{subfigure}
     \hfill
     \begin{subfigure}{0.23\linewidth}
         \centering
         \includegraphics[width=\linewidth]{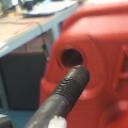}
         \caption{Insert(R)}
         \label{img_insert1}
     \end{subfigure}
     \hfill
     \begin{subfigure}{0.23\linewidth}
         \centering
         \includegraphics[width=\linewidth]{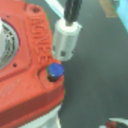}
         \caption{Screw(L)}
         \label{img_screw1}
     \end{subfigure}
     \hfill
     \begin{subfigure}{0.23\linewidth}
         \centering
         \includegraphics[width=\linewidth]{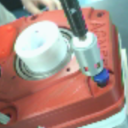}
         \caption{Screw(R)}
         \label{img_screw2}
     \end{subfigure}
     \hfill
        \caption{Binocular Camera View. L, left camera view. R, right camera view. Pick and Place are omitted. Agent perform the task with this as observation.}
        \label{BinocularObservationInRealTask}
\end{figure}

We formulate this visual servoing process as a partially observable Markov decision process (POMDP). \textbf{Action space} \(\overrightarrow{a} \in A^x \times A^y\) (robot's base Cartesian frame) is given by the position control for the controlled point \(P_A\). This process is rolling out under discrete time step \(t\in\left[1;T\right]\), \(T\) is the max steps in each episode. Observation space is giving by two low-resolution RGB images. The agent detects the controlled point \(P_A\) and target point \(P_B\) in the raw images as the \textbf{observation space}. \textbf{Reward} \(r_t\) is a scalar variable only given in simulation. Transition process is described as \(p\left(o_t,r_t|a_{t-1},o_{t-1}\right)\). The agent in the simulation should be trained to maximize the total rewards \(\sum r_t\) in the whole process.

\begin{figure*}
    \centering
    \includegraphics[width=\textwidth]{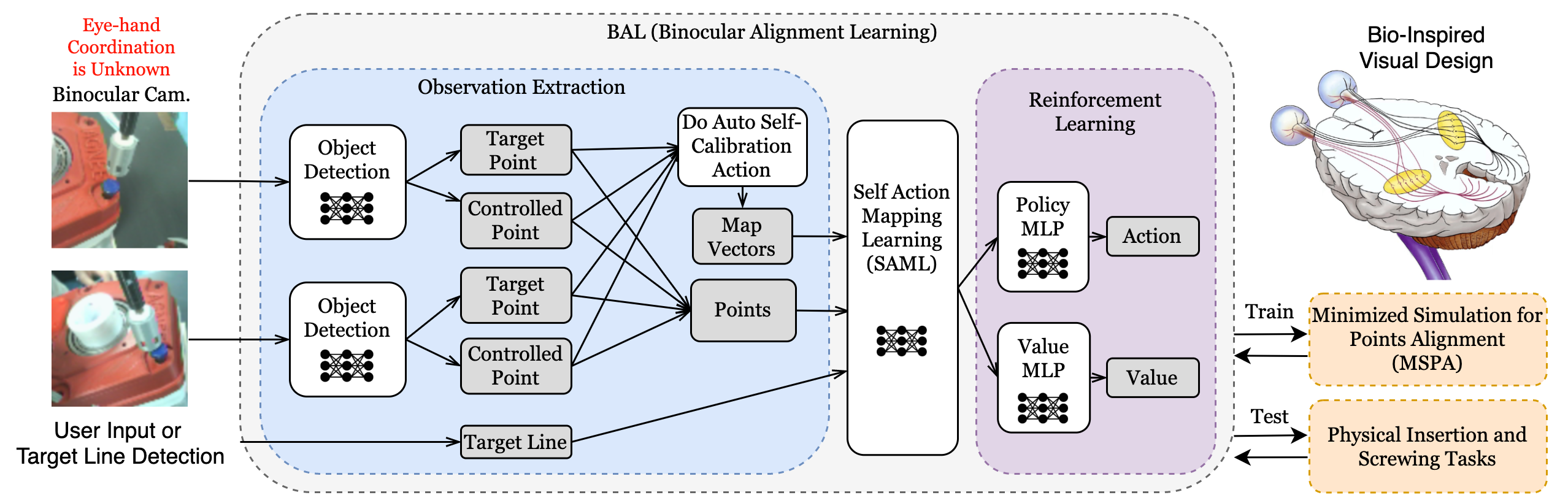}
    \caption{Binocular Alignment Learning (BAL) Architecture. A pre-trained YOLO network is used to detect key points. Details about auto self-calibration action can refer to Sec.\ref{section_self_mapping}. Target line detection will be discussed in future work, this work we use user input target line.}
    \label{Fig_ModelArchitec}
\end{figure*}
\section{Approach}

\begin{figure}
    \centering
    \includegraphics[width=0.6\linewidth]{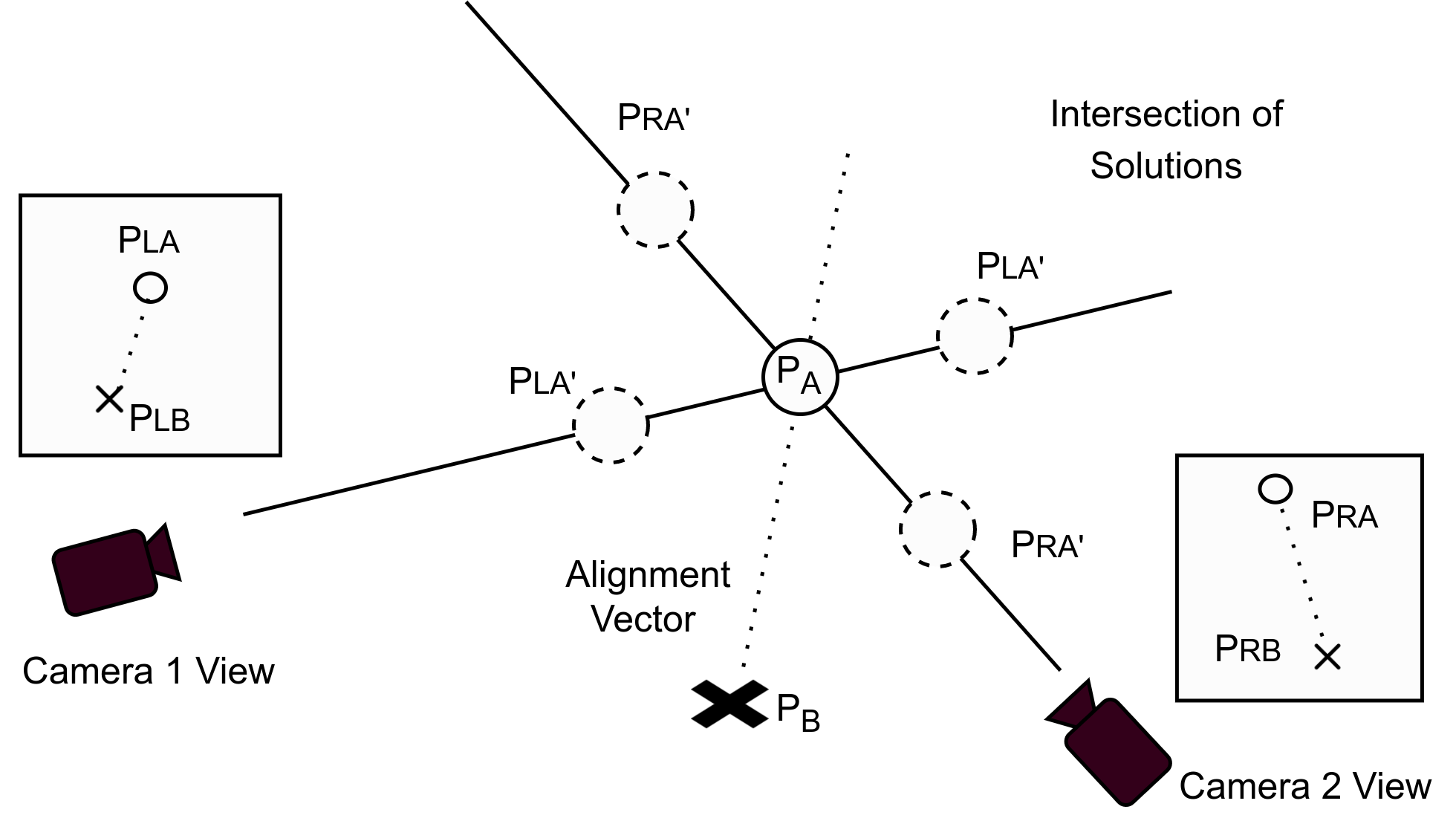}
    \caption{An illustration of the intersection of solutions in Binocular Vision}
    \label{IntersectionofSolutions}
\end{figure}

\subsection{Approach Overview}

\begin{figure}
    \centering
    \includegraphics[width=0.8\linewidth]{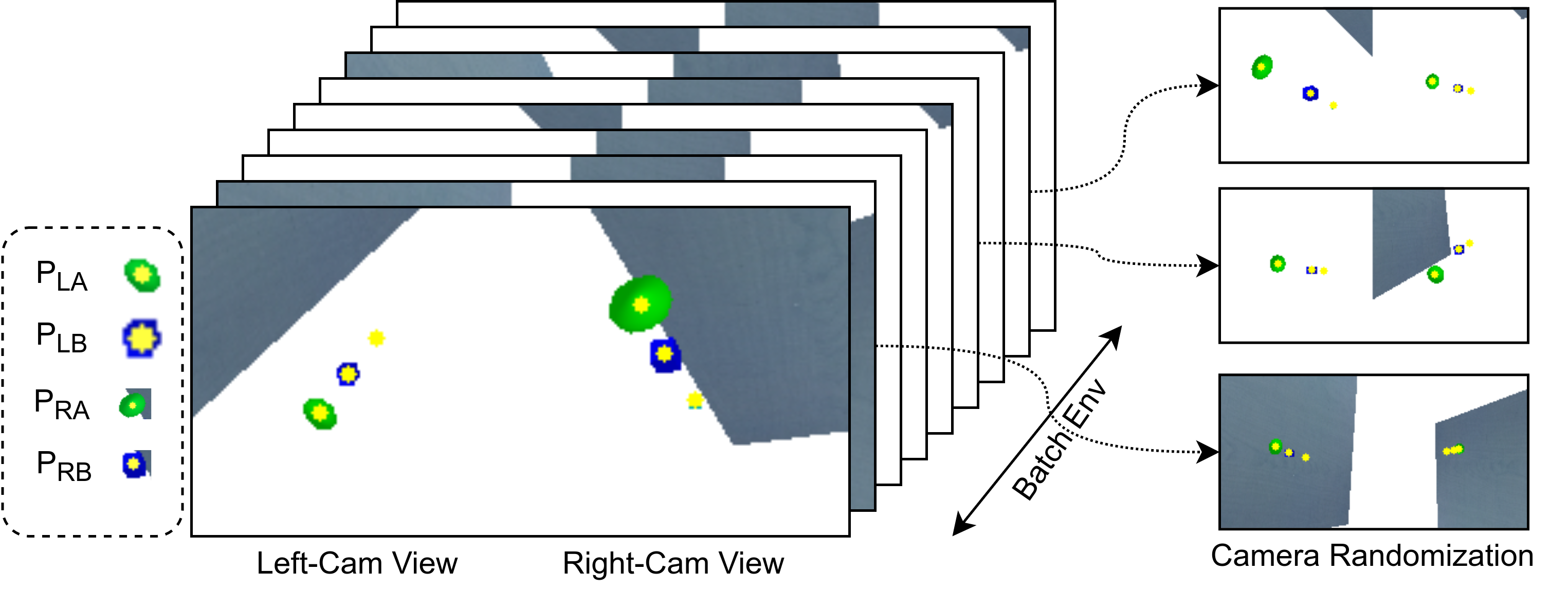}
    \caption{This is the Minimal simulation for Points Alignment (MSPA), which only provides points to train policy.}
    \label{Fig_MSPA}
\end{figure}

\begin{figure}
    \centering
    \includegraphics[width=\linewidth]{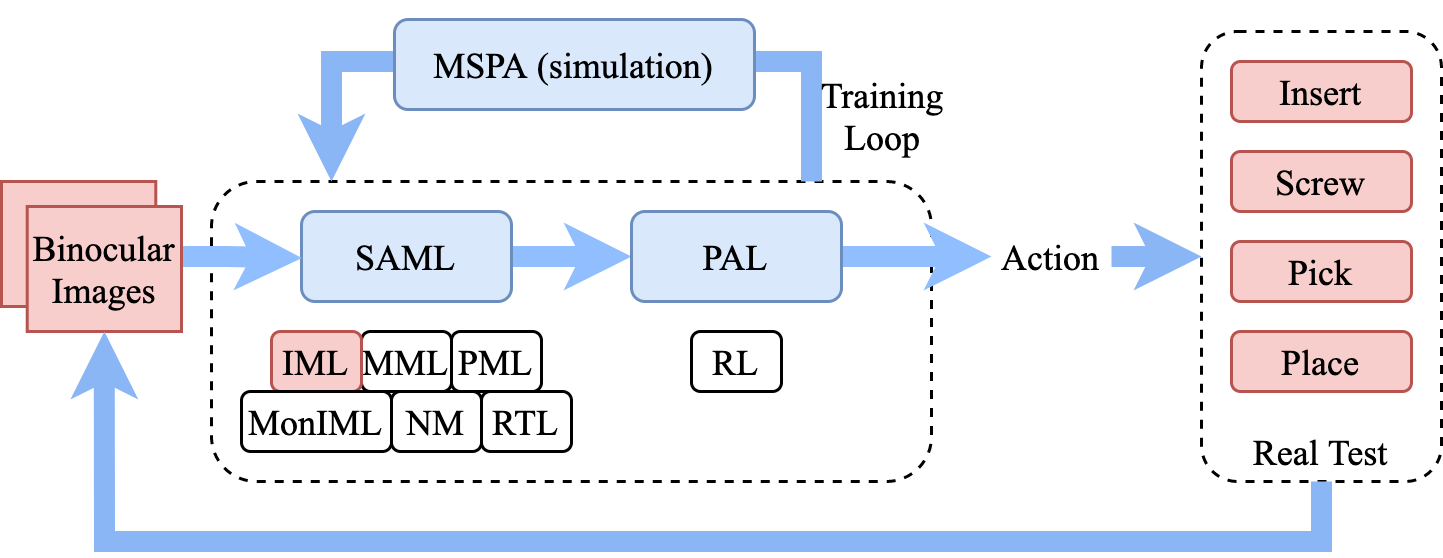}
    \caption{Components Workflow. SAML:Self Action Mapping Learning (in Sec.\ref{section_self_mapping}); PAL:Points Alignment Learning (in Sec.\ref{sec_pal}); MSPA: Minimal Simulation for Points Alignment (in Sec.\ref{sec_intro_of_simulation})}.
    \label{Fig_components_workflow}
\end{figure}

This learning part introduces the Self Action Mapping Learning (SAML) to correct eye-hand coordination and Points Alignment Learning (PAL) to perform task policy. Minimal Simulation for Points Alignment (MSPA, Sec.\ref{sec_intro_of_simulation} and Fig.\ref{Fig_MSPA}) is used to efficiently learn the policy in a few minutes training. The general approach is described in Fig.\ref{Fig_ModelArchitec}, Fig.\ref{Fig_components_workflow} and Algorithm.\ref{alg:two}. Fig.\ref{Figure_self_action_map_learning} gives network designs for SAML methods. 

The inputs to the model are two images. The model is allowed to use the object detector (we use YOLO) to capture the key points in the images, annotated as \(P_{ij} \in \{P_{LA},P_{LB},P_{RA},P_{RB}\}\), \(L,R\) stand for the left and right cameras selection, \(A,B\) represents the controllable point and target point, seeing Fig.\ref{IntersectionofSolutions}. Using the self action mapping learning (SAML) methods, the robot generates the self-calibration vector (SCV) \(V_{ik}\), \(i\in\{L,R\}\) and \(k \in \{1,2\}\).Using the SCV and \(P_{ij}\), the model learns a camera-pose adaptive control policy using Self Action Mapping Learning (SAML). While testing on the real robot, the RL agent and YOLO works at $1\sim2Hz$, the robot controlled by ROS Moveit at $20Hz$ level, camera $30Hz$.

\RestyleAlgo{ruled}

\SetKwComment{Comment}{/* }{ */}
\SetKwInput{KwInit}{Init}
\SetKwInput{KwInput}{Input}
\SetKwBlock{Train}{Train in Simulation (MSPA)}{end}
\SetKwBlock{Test}{Test in Real}{end}
\SetKw{KwTrain}{Train(Sim)}
\begin{algorithm}[hbt!]
\caption{Binocular Alignment Learning and Sim2Real}\label{alg:two}
\KwInit{RL Agent $\Psi(obs)$, Trained Object Detector $Y(Image)$, min error $\varepsilon$}
\Train{ 

    \While{Train}{
        \ForEach{Episode}{
        
            $P_A, P_{camera}$ $\gets$ Random()\;
            $env\gets$CreateEnv($P_A, P_{camera}$)\;
            Interact($\pi_\Psi,env$) until done \;
        }
        \If{Update}{Gradient Update $\Psi$ (PPO Convention)}
    }
}

\Test{
    \KwInput{Target Alignment Line $\overrightarrow{H}$}
    $SCV_{1,2}$ $\gets$ do auto calibration action, collect\;
    
    \Comment{Alignment}
    \While{err$>\varepsilon$}{
    $P_{i}$ $\gets$ Y($Images$)\;
            obs$\gets$Preprocess$(P_{i},\overrightarrow{H},SCV)$\;
        
        $a\gets\pi_{\Psi}(obs)$\;
        
        robot input ($a$)\;
        
        err$\gets$GetError($\overrightarrow{H},P_{i}$)\;
    }
        
        
        
        
    
    \Comment{Alignment Finished}
    Do action $\in$ $\{insert, screw, pick, place\}$\;
    
}

    
        
        
        
        
    
    

\end{algorithm}

\subsection{Self Action Mapping Learning (SAML)\label{section_self_mapping}}

    \begin{figure}
        \centering
        \includegraphics[width=\linewidth]{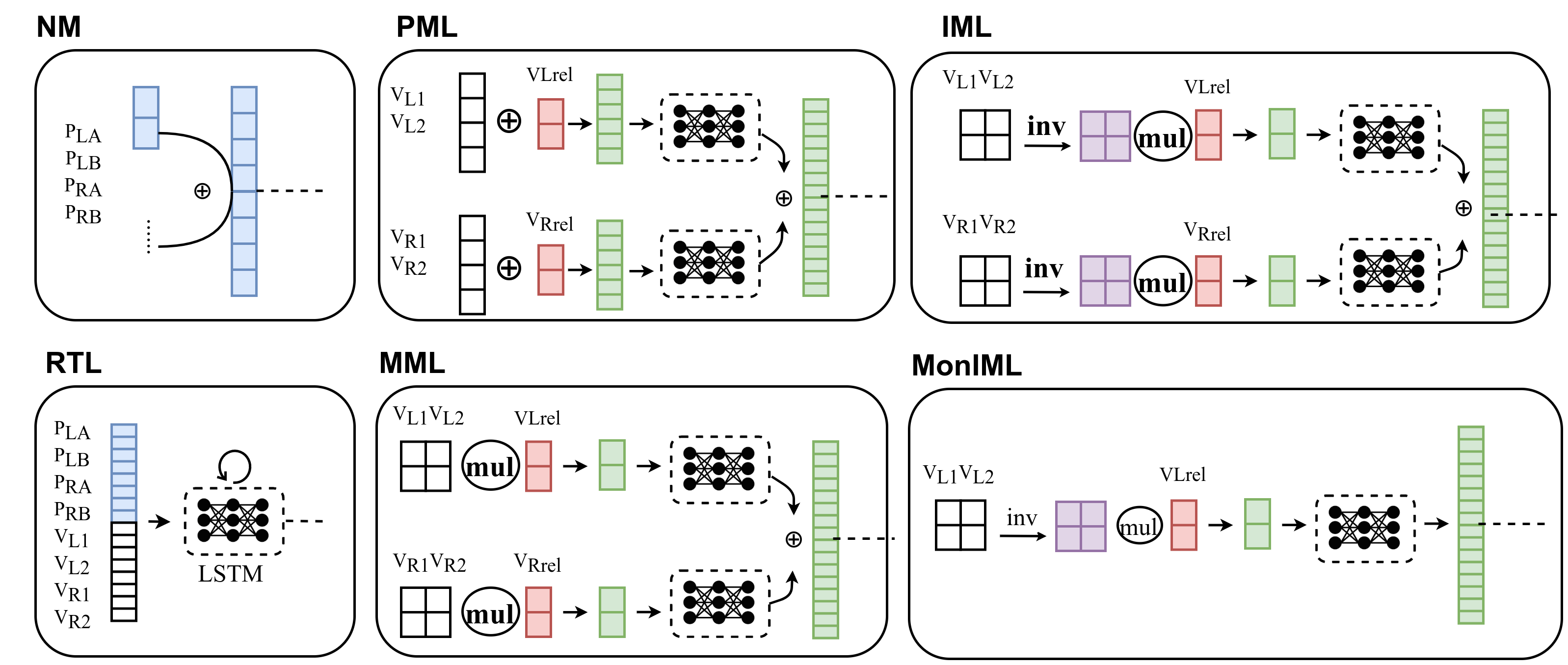}
        \caption{Self Action Mapping Learning (SAML) methods }
        \label{Figure_self_action_map_learning}
    \end{figure}

To address the weak eye-hand coordination problem, we propose methods to achieve the self action mapping learning. We designed the PML, MML, IML and MonoIML. In those approaches the agent need to perform a self-calibration action \(a_{d1}=\left[1,0\right]\) (move towards \(A_x\) direction for 1 unit), come back to initial position and perform 
\(a_{d2}=\left[0,1\right]\) (move towards \(A_y\) direction for 1 unit) 
in sequence to collect the corresponding position changes in the observation. They are annotated as self-calibration vector (SCV) \(V_{ik}\) with 
\(i \in \{ L,R\}, k \in \{1,2\}\)
representing the position translation of \(P_{iA}\) as the result of action \(a_{d1}\) and \(a_{d2}\). Target related vector (TRV) \( V_{irel} \) represents the relative position in the camera observation of \(P_{iA},P_{iB}, i \in \{L,R\}\). \(L_\Theta\) is the MLP (multilayer perceptron) block for information extraction out of vectors. \(F\) is the flatten layer. Given \(H_{L},H_{R}\) as the alignment target vector by user. All the pipelines are described in the Fig.\ref{Figure_self_action_map_learning}.
\[V_{L1},V_{R1} \hookleftarrow a_{d1}\]
\[V_{L2},V_{R2} \hookleftarrow a_{d2}\]
\[V_{Lrel},V_{Rrel}  = V_{P_{LA}P_{LB}}-H_{L},V_{P_{RA}P_{LB}}-H_{R}\]



\textbf{None-Maping (NM)} has no action-mapping learning. Hence the robot only observe the object detection results from the last layer. With the random noise given to the camera position, this approach should perform the worst. This approach serves as baselines to be compared. 
\textbf{Monocular Mapping Learning (MonIML)} utilizes only one camera observation in IML. $o = L_\Theta (matmul((V_{i1};V_{i2})^{-1},V_{irel})), i \in \{L\}$


\textbf{Parral Mapping Learning (PML)} concatenates SCV with TRV and flattens them into a 1D tensor. $h_i = L_\Theta ((V_{i1};V_{i2}),V_{irel}), i \in \{L,R\}$;$o = F(h_L;h_R)$ However, since SCV and TRV are from a different domain, it can be difficult for the model to learn the relation between SCV and TRV. 

\textbf{Recurrent Time-based Learning (RTL)} recurrently process TRV, $h,o=LSTM(h,[V_{lrel};V_{rrel}])$ to learn the action-observation coordination.

\textbf{Mat-mul Mapping Learning (MML)} multiplies the SCV and TRV, $h_i = L_\Theta (matmul((V_{i1};V_{i2}),V_{irel})), i \in \{L,R\}$; $o = [h_L;h_R]$. 

\textbf{Inverse Mapping Learning (IML)} multiplies the inverse of SCV (\(V_{ik}\))to the TRV. It inverses the SCV into an easy learning domain \(U\) (details of experiments are in Sec.\ref{sec_experiments}). $h_i = L_\Theta (matmul((V_{i1};V_{i2})^{-1},V_{irel})), i \in \{L,R\}$; $o = [h_L;h_R]$

\subsection{Points Alignment Learning (PAL)\label{sec_pal} and Minimal Simulation  (MSPA)\label{sec_intro_of_simulation}}

Points Alignment task learning is described as a Markov Decision optimization problem introduced in Sec.\ref{sec_taskdef}. The Proximal Policy Optimization (PPO) approach is used to train the policy for action generation. The rewards are defined as \(r=clip\left(-80 * D + 1, -10, 1\right)\) to help the agent learn to approach the perform alignment task. \(D\) is the distance between the current controllable point position and target position. \(D\) is only applicable in the simulation, while in the real inference there is no \(D\).
In the simulation, seeing Fig.\ref{Fig_MSPA}, the camera position is randomized to help learn a camera position adaptive strategy. The key points are given in the simulation as the green point \(P_B\) and the blue point \(P_A\). The third yellow point is a random dot located on the given target alignment vector \(H\) (Sec.\ref{sec_taskdef}). For learning-based methods, training in the simulation then testing in the physical world is much more efficient and less dangerous than directly training in the physical environment. With feature disentanglement and domain randomization, RL policy optimization (PPO convention) the policy can successfully adapt to the real test.

\section{Experiment and Discussion\label{sec_experiments}}

\begin{figure}
     \centering
     \begin{subfigure}[b]{0.48\linewidth}
         \centering
         \includegraphics[width=0.8\linewidth]{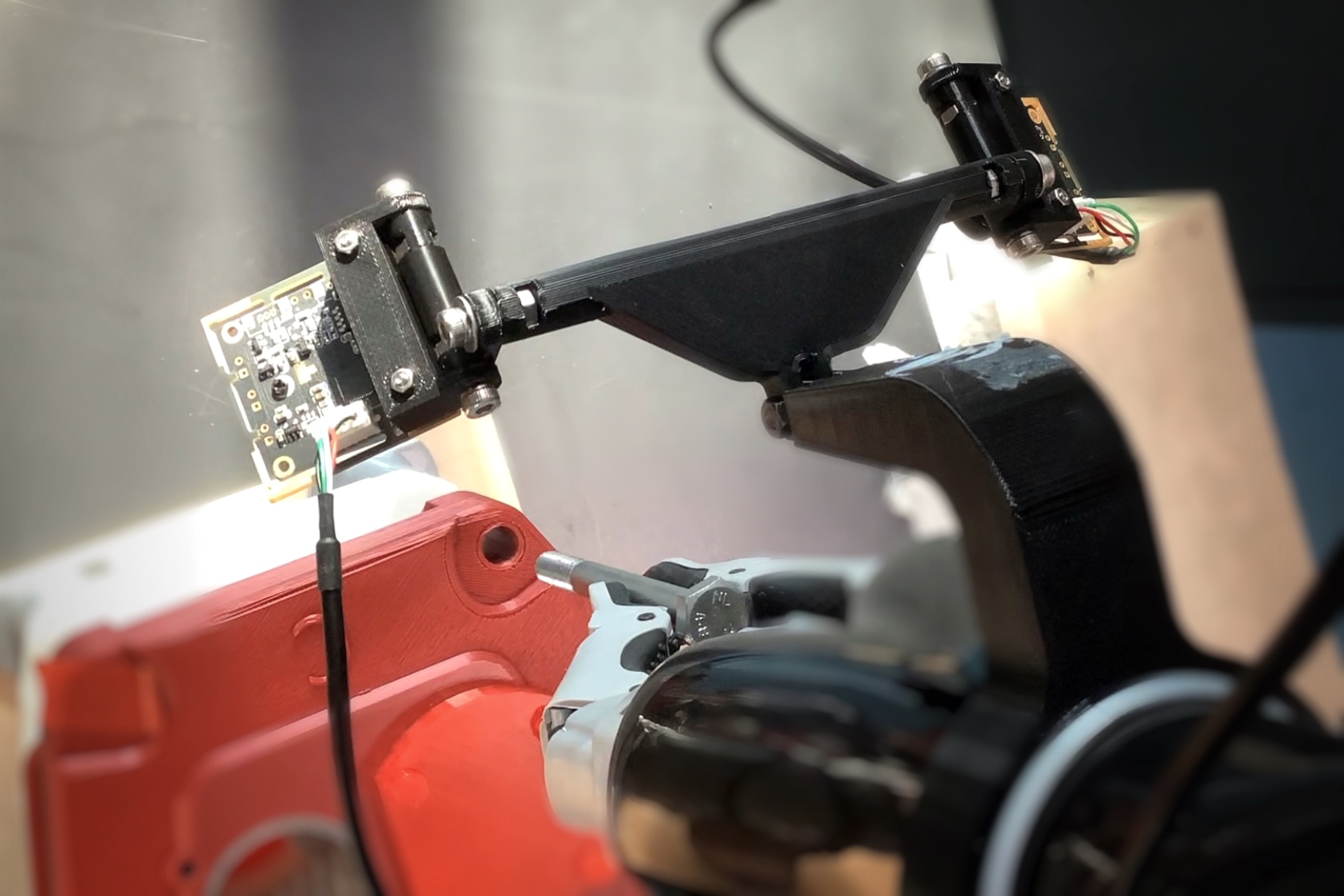}
         \caption{Insert}
         \label{image_task_setup_physical_ins}
     \end{subfigure}
     \hfill
     \begin{subfigure}[b]{0.48\linewidth}
         \centering
         \includegraphics[width=0.8\linewidth]{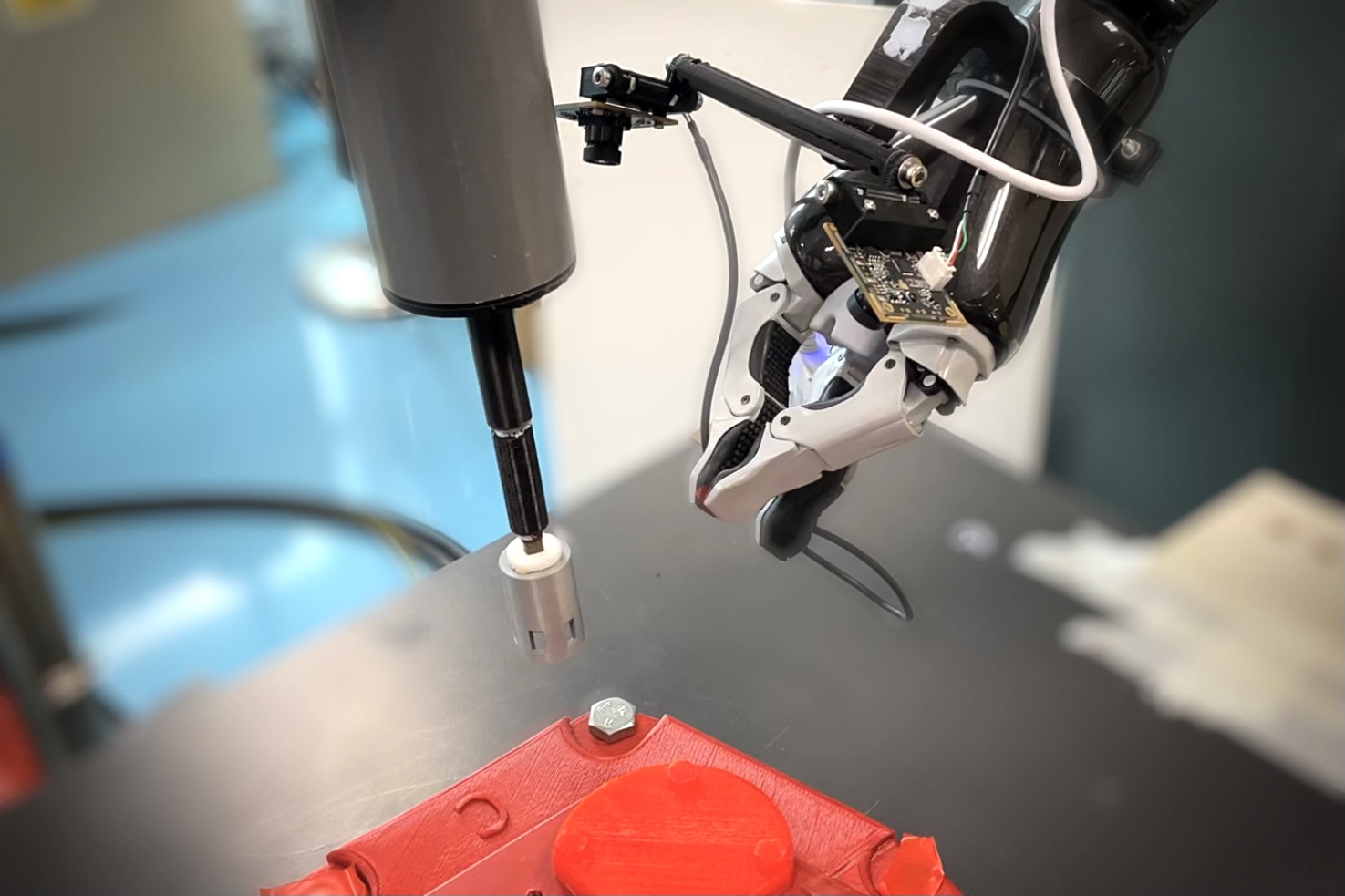}
         \caption{Screw.}
         \label{image_task_setup_physical_screw}
     \end{subfigure}
     
      \begin{subfigure}[b]{0.48\linewidth}
         \centering
         \includegraphics[width=0.8\linewidth]{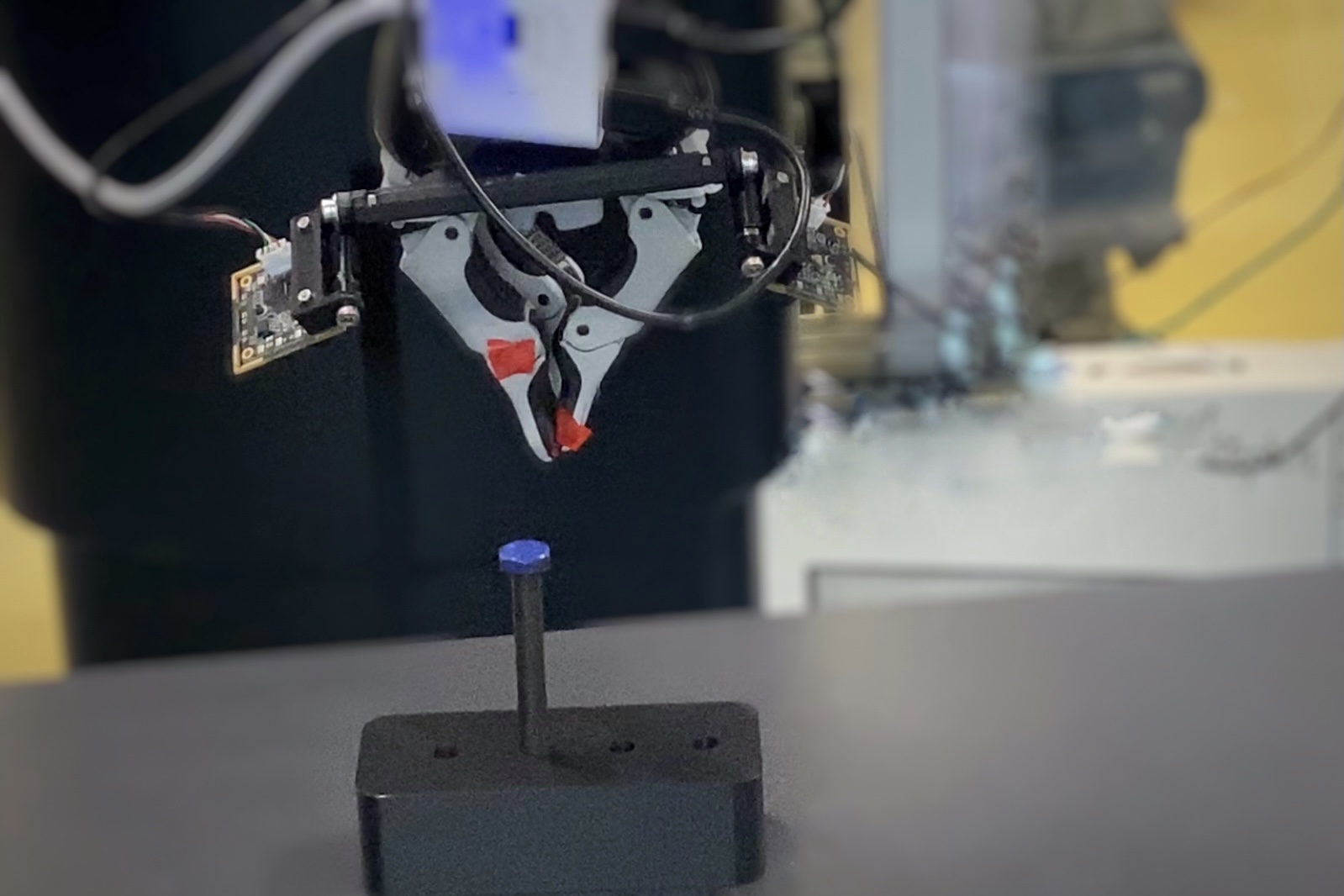}
         \caption{Pick-small-item}
         \label{image_task_setup_physical_pick}
     \end{subfigure}
     \hfill
     \begin{subfigure}[b]{0.48\linewidth}
         \centering
         \includegraphics[width=0.8\linewidth]{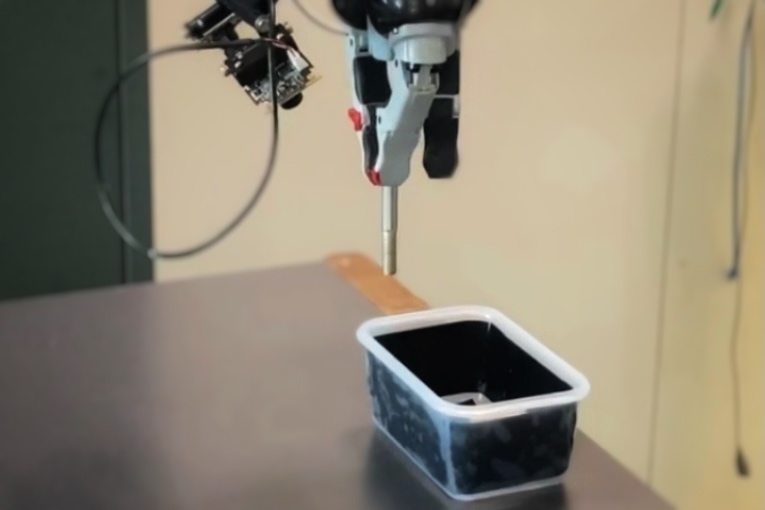}
         \caption{Place-small-box}
         \label{image_task_setup_physical_place}
     \end{subfigure}
     
        \caption{Task setup. We use dual arm KINOVA MOVO robot, and the binocular camera is set on the last joint of the right arm. *For the screwing task, the right gripper holds the camera, and the left manipulator perform the task. The camera needs no eye-hand calibration.}
        \label{image_task_setup_physical}
\end{figure}

\begin{table}
\centering
\caption{Ablation study of Self Action Mapping Learning (SAML) and Camera Randomized Training. FC use fixed camera pose. RC use randomized camera pose. Different methods are introduced in Sec.\ref{section_self_mapping}. \label{table_saml_test_result}. This result is collected from simulation.}
\begin{tabular}{c cc cc}
\hline
   Sim. Test     & \begin{tabular}[c]{@{}l@{}}FC Train\\ FC Test\end{tabular} & \begin{tabular}[c]{@{}l@{}}FC Train\\ RC Test\end{tabular} & \textbf{\begin{tabular}[c]{@{}l@{}}RC Train\\ FC Test\end{tabular}} & \begin{tabular}[c]{@{}l@{}}RC Train\\ RC Test\end{tabular} \\ \hline
PML          & 2\%                                                       & 9\%                                                       & 4\%                                                                & 20\%                                                       \\
MML          & 3\%                                                      & 0\%                                                          & 91\%                                                       & 59\%                                                       \\
NM           & 6\%                                                       & 6\%                                                       & 1\%                                                                & 11\%                                                       \\
MonIML       & 3\%                                                       & 15\%                                                       & 36\%                                                                & 26\%                                                       \\
RTL          & 2\%                                                       & 4\%                                                       & 3\%                                                                & 61\%                                                       \\
\textbf{IML} & \textbf{11\%}                                              & \textbf{37\%}                                              & \textbf{96\%}                                                       & \textbf{66\%}   \\  \hline

\end{tabular}%

\end{table}

We conduct experiments to answer to the following questions:
\textbf{Binocular Vision:} The advantage using binocular vision rather than monocular vision (in Sec.\ref{sec_exp_mono_or_bino}).
\textbf{Eye-hand Calibration:} Can BAL successfully learn the camera pose adaptive policy? i.e., solve task under a poor eye-hand calibration. (in Sec. \ref{sec_exps_eyehand_cali}).
\textbf{Human Performance:} Using the same setup, how does human perform on this task (in Sec.\ref{exp_limitation_and_future_work})?

We conduct experiments in both simulation and real robots. As for the physical experiment, we set up the robot to perform four Points Alignment tasks insert, screw, pick-small-item, and place-small-box, seeing the setup in Fig.\ref{image_task_setup_physical}. Four tasks are in the domain of Points Alignment. Furthermore, Points Alignment can also be used in box stacking, pushing, pouring water into the cup and,.etc.

\subsection{Benchmark and Eye-hand Coordination Learning\label{sec_exps_eyehand_cali}}

We compared our work in Table.\ref{Benchmark} with recent years' learning-based methods in high precision tasks, especially in the task of insertion. BAL performs competitively without camera calibration and with minimal infrastructure, only two low-cost cameras. In contrast, others are much more complex and expensive. BAL also shows robustness to a large work range of more than 50mm comparing with others less than 40mm or 2mm.

Ablation study of different SAML methods is given in the Table.\ref{table_saml_test_result}. To overcome weak eye-hand calibration, BAL(IML) has the best success rate of $96\%$ with a camera position adaptive policy. Random camera poses training benefits from domain randomization and improves the performance from $11\%$ to $96\%$. It also shows fixed camera training can not solve an unseen eye-hand coordination situation. Table.\ref{Success Rate on real robot} shows BAL(IML) can solve tasks with a success rate of $91.9\%-100\%$ in a real robot. Results also show using a inverse method, the IML improved the performce from MML(91\%) to IML(96\%).

\subsection{Binocular is better than Monocular\label{sec_exp_mono_or_bino}}

To show the necessity of a binocular camera system compared with a monocular camera, we give a baseline using MonIML (Monocular-based IML). Table.\ref{table_saml_test_result} shows, under RC training and RC testing setup, IML (96\%) successes much more than MonIML (36\%). Using FC training, IML (11\% in FC test and 37\% in RC test) also shows much better performance than MonIML (3\% in FC test and 15\% in RC test). Therefore, binocular-based methods (i.e., IML) learns a better policy in solving target tasks, and the policy is adaptive to camera position.

\begin{table*}[]\small
\centering
\caption{\centering Points Alignment Tasks Benchmark Comparison. Manual Calib. indicates if this method requires manual eye-hand coordination calibration. N, not applicable or not given. Vis., Vision-based Method. Est. Cost, Estimation of Cost.} 
\resizebox{\textwidth}{!}{
\begin{tabular}{c ccccccc}
\hline
Method               & Vis.                      & Manual Calib.        & Est. Cost (\$) & Success Rate          & Max Range (mm)              & Additional Sensors \& Info     &  \\ \cline{1-7}
M. V. et al, \cite{1_practial_insert}, 2019 & \checkmark & Need                 & 510            & 97\%                  & \textless{}40             & RGB Cam., T/F Sensor           &  \\
G. T. et al, \cite{8_assemblycad}, 2018 & \checkmark & Need                 & 350            & \textbf{100\%}                   & N                         & 3D Cam., Goal Pose             &  \\
F. B. et al, \cite{7_learning_to_touch}, 2018 & \checkmark & No Need              & \textbf{20}             & N                     & N                         & 2×RGB Cam                      &  \\
J. C. et al, \cite{6_quick_insert}, 2019 & \checkmark & Need                 & 520            & \textbf{100\%}                 & N                         & 2×RGB Cam., T/F Sensor         &  \\
G. S. et al, \cite{5_rl_insert}, 2019 & \checkmark & Need                 & 510            & \textbf{84\%-100\%}            & \textless{}1              & RGB Cam., Goal Pose Image, T/F &  \\
T. I., \cite{11prcesision_assembly}, 2017      & \XSolid     & N                    & 500            & \textbf{100\%}                 & \textless{}2              & T/F Sensor                     &  \\
T. T. et al, \cite{3_learn_peg_hole_insert}, 2015 & \XSolid     & N                    & 500            & 86.7\%-90\%           & 0                         & T/F Sensor                     &  \\
B. C. et al, \cite{2_pegoHoleInsert}, 2020 & \XSolid     & N                    & 500            & \textbf{65\%-100\%}            & \textless{}4              & T/F Sensor, Goal Pose          &  \\ 
Human (BAL Setup)   & \checkmark & N              & N              & 20\%-40\%             & \textgreater{}50          & N                              &  \\ 
\textbf{BAL(Ours)}                & \checkmark & \textbf{Weakly Need} & \textbf{20}    & \textbf{91.9\%-100\%} & \textbf{\textgreater{}50} & 2×RGB Cam., Target Line            &  \\ 
\hline
\end{tabular}
}

\label{Benchmark}
\end{table*}

\begin{table}[]\small

\centering
\caption{Success Rate on Real Robot using policy BAL+IML. Test setup is shown in Fig.\ref{image_task_setup_physical} \label{success rate}}
\begin{center}

\begin{tabular}{c cccc}
\hline
Real Test         & Insert         & Screw          & \begin{tabular}[c]{@{}l@{}}Pick\\ (small-item)\end{tabular} & \begin{tabular}[c]{@{}l@{}}Place\\ (small-box)\end{tabular} \\ \hline

Success Rate & $91.9\pm1.9\%$ & $93.7\pm6.3\%$ & $100\%$  & $100\%$                          \\ \hline                          
\end{tabular}%

\end{center}

\label{Success Rate on real robot}
\end{table}

\subsection{Discussions and Future Work\label{exp_limitation_and_future_work}}

\begin{figure}
    \centering
    \includegraphics[width=0.6\linewidth]{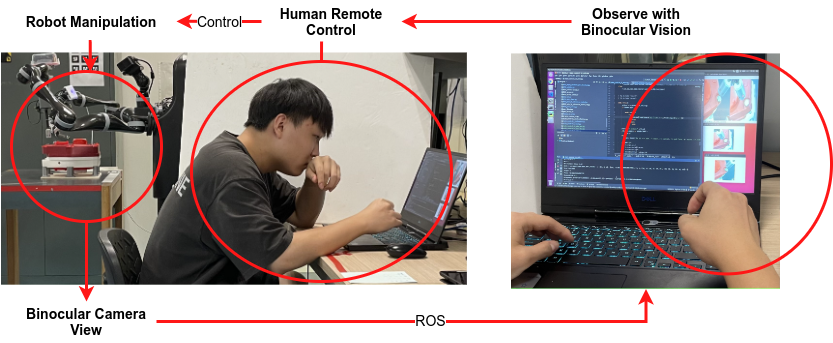}
    \caption{Human Performance Test. The player is only allowed to observe two images \(I_L, I_R\) given by camera and should input with the robot action \(a\) to remote control the robot. The eye-hand coordination is unknown to the player.}
    \label{Fig_human_performance}
\end{figure}

\textbf{Human Performance:} We also set up the experiment compared with human performance in Fig.\ref{Fig_human_performance}. With the unknown eye-hand calibration, he needs to learn it from trial and error, as what the agent will do in the Sec.\ref{section_self_mapping}. Interestingly and beyond our expectation, the human perform much worse than the agent achieves only $30\%\pm10\%$ success rate (10 attempts) in insertion. \textbf{Recurrent Policy:} Our physical test results are given using IML. The reason for not using RTL is that it can randomly cause the controllable point to go out of the camera view. However, it easily escaped from the camera view and invalidated the policy. In the application of manufacturing, we need the method to show consistency among experiments.
\textbf{Limitations and Future Work:} However, there is still some limitation about our work, and we will address it in future work. \textbf{1.}We didn't discuss the camera distortion and give an ablation study on the camera distortion with SAML. \textbf{2.}This version of our work reduces the efforts to calibrate the camera but still needs humans to intervene to give the target line. In the future, we want the robot to auto-detect the target alignment line. \textbf{3.}As a pure visual-based method in such a big work range, we also haven't reach 100\% in insertion, just like a lot of other methods, we will further research in this area and improve the performance to 100\%. \textbf{4.}We will also explore more binocular applications and verify them on more precise tasks.

\section{Conclusion}

Precision manipulation is a long-term challenge in robotics. In this work, we propose BAL to successfully solve insertion, screwing, pick-small-item, and place-small-box with success rate of $91.9\%-100\%$. Additionally, we also reduced the cost of the setup, making it economically efficient. We addressed the importance of adaptability under poor eye-hand coordination and proposed SAML methods to solve it with a detailed ablation study. We proposed a practical sim2real pipeline and successfully adapt it to real robot test.

%

%
%


\begin{thebibliography}{5}
%
\bibitem{11prcesision_assembly}
T.  Inoue,  G.  De  Magistris,  A.  Munawar,  T.  Yokoya, and  R.  Tachibana,  “Deep  reinforcement  learning  for high  precision  assembly  tasks,”  in \emph{2017 IEEE/RSJInternational Conference on Intelligent Robots and Systems (IROS)}, 2017, pp. 819–825.

\bibitem{2_pegoHoleInsert}
C. C. Beltran-Hernandez,  D. Petit,  I. G. Ramirez-Alpizar, and K. Harada, “Variable  compliance  control   for   robotic   peg-in-hole   assembly:   A   deep-reinforcement-learning  approach,”Applied Sciences,vol. 10, no. 19, 2020, \scriptsize{ISSN}: \small{2076-3417.}

\bibitem{Fang_2020_CVPR_graspnet}
H.-S. Fang, C. Wang, M. Gou, and C. Lu, “Graspnet-1billion:  A  large-scale  benchmark  for  general  object grasping,” in \emph{CVPR}, Jun. 2020.

\bibitem{zeng2019tossingbot}
A.   Zeng,   S.   Song,   J.   Lee,   A.   Rodriguez,   and   T. Funkhouser, “Tossingbot: Learning to throw arbitraryobjects with residual physics,” 2019.

\bibitem{goodrobot_topdownview_grasp}
A.  Hundt,  B.  Killeen,  H.  Kwon,  C.  Paxton,  and  G.Hager,  “”good  robot!”:  Efficient  reinforcement  learn-ing  for  multi-step  visual  tasks  via  reward  shaping,”Sep. 2019.

\bibitem{12metarl_insert}
G.  Schoettler,  A.  Nair,  J.  A.  Ojea,  S.  Levine,  and E. Solowjow, \emph{Meta-reinforcement learning for robotic industrial insertion tasks}, 2020.

\bibitem{graspnet}
A. Mousavian, C. Eppner, and D. Fox, “6-dof graspnet: Variational grasp generation for object manipulation, ”Oct. 2019, pp. 2901–2910. \scriptsize{DOI}\small{: 10.1109/ICCV.2019.00299.}

\bibitem{eye-hand-cali}
H.  Cheng,  Z.  Zhang,  and  W.  Li,  “Efficient  hand  eyecalibration  method  for  a  delta  robot  pick-and-place system,”  in \emph{2015 IEEE CYBER},  2015,  pp.  175–180. \scriptsize{DOI}\small{: 10.1109/CYBER.2015.7287930.}

\bibitem{5_rl_insert}
 G. Schoettler, A. Nair, J. Luo, S. Bahl, J. A. Ojea, E.Solowjow, and S. Levine, \emph{Deep reinforcement learning for industrial insertion tasks with visual inputs and natural rewards}, 2019.

\bibitem{7_learning_to_touch}
F.  de  La  Bourdonnaye,  C.  Teuli`ere,  J.  Triesch,  andT. Chateau, “Learning to touch objects through stage-wise deep reinforcement learning,” in \emph{2018 IEEE/RSJ International Conference on Intelligent Robots and Systems (IROS)},  2018,  pp.  1–9. \scriptsize{DOI:}\small{  10.1109/IROS.2018.8593362.}

\bibitem{3_learn_peg_hole_insert}
\emph{A Learning-Based Framework for Robot Peg-Hole-Insertion}, Dynamic Systems and Control Conference, V002T27A002, Oct. 2015. \scriptsize{DOI:} \small{10.1115/DSCC2015-9703.}

\bibitem{1_practial_insert}
M. Vecerik, O. Sushkov, D. Barker, T. Rothorl, T. Hes-ter,  and  J.  Scholz,  “A  practical  approach  to  insertion with  variable  socket  position  using  deep  reinforcement  learning,”  in \emph{2019 International Conference on Robotics and Automation (ICRA)}, 2019, pp. 754–760. \scriptsize{DOI:} \small{10.1109/ICRA.2019.8794074.}

\bibitem{8_assemblycad}
G.  Thomas,  M.  Chien,  A.  Tamar,  J.  A.  Ojea,  and P.  Abbeel,  “Learning  robotic  assembly  from  cad,”  in \emph{2018 IEEE International Conference on Robotics and Automation (ICRA)},  2018,  pp.  3524–3531. \scriptsize{DOI:}\small{  10 .1109/ICRA.2018.8460696.}

\bibitem{openai2019solvingcubic}
OpenAI, I. Akkaya, M. Andrychowicz, M. Chociej, M.Litwin, B. McGrew, A. Petron, A. Paino, M. Plappert,G. Powell, R. Ribas, J. Schneider, N. Tezak, J. Tworek,P.  Welinder,  L.  Weng,  Q.  Yuan,  W.  Zaremba,  and  L.Zhang, \emph{Solving rubik’s cube with a robot hand}, 2019.

\bibitem{ppopaper}
 J. Schulman, F. Wolski, P. Dhariwal, A. Radford, andO.  Klimov, \emph{Proximal policy optimization algorithms},2017.


\bibitem{6_quick_insert}
 J.  Triyonoputro,  W.  Wan,  and  K.  Harada, \emph{Quickly inserting pegs into uncertain holes using multi-view images and deep network trained on synthetic data}, Feb. 2019.








\end{thebibliography}
\end{document}